\documentclass[runningheads, envcountsame, a4paper]{llncs}

\usepackage{amsmath}
\usepackage{amsthm}
\usepackage{booktabs}
\usepackage{amsfonts}
\usepackage{algorithm}
\usepackage{algorithmic}
\usepackage{subfigure}
\usepackage{siunitx}
\usepackage{comment}
\usepackage{arydshln}
\usepackage[pdftex]{graphicx}
\usepackage{epstopdf}
\usepackage[hyphens]{url}
\usepackage{microtype}
\usepackage[misc]{ifsym}
\newcommand{\tabincell}[2]{\begin{tabular}{@{}#1@{}}#2\end{tabular}}

\begin{document}
\title{Graph-Revised Convolutional Network}
\toctitle{Graph-Revised Convolutional Network}

%
\author{Donghan Yu \Letter \and Ruohong Zhang\and Zhengbao Jiang \and \\ Yuexin Wu \and Yiming Yang}
\tocauthor{Donghan Yu (Carnegie Mellon University),
Ruohong Zhang (Carnegie Mellon University),
Zhengbao Jiang (Carnegie Mellon University),
Yuexin Wu (Carnegie Mellon University),
Yiming Yang (Carnegie Mellon University)
}

\authorrunning{D. Yu et al.}
%
\institute{Carnegie Mellon University, Pittsburgh PA 15213, USA \\
\email{$\{$dyu2,ruohongz,zhengbaj,yuexinw,yiming$\}$@cs.cmu.edu}}
\maketitle              

\newcommand{\Model}{Graph-Revised Convolutional Network }
\newcommand{\Modelshort}{GRCN }
\newcommand{\Modelshortbr}{(GRCN)}

\begin{abstract}
Graph  Convolutional  Networks  (GCNs)  have  received  increasing  attention in the machine learning community for effectively leveraging both the content features of nodes and the linkage patterns across graphs in various applications. As real-world graphs are often incomplete and noisy, treating them as ground-truth information, which is a common practice in most GCNs, unavoidably leads to sub-optimal solutions. Existing efforts for addressing this problem either involve an over-parameterized model which is difficult to scale, or simply re-weight observed edges without dealing with the missing-edge issue. This paper proposes a novel framework called Graph-Revised Convolutional Network \Modelshortbr, which avoids both extremes. 
Specifically, a GCN-based graph revision module is introduced for predicting missing edges and revising edge weights w.r.t. downstream tasks via joint optimization. A theoretical analysis reveals the connection between \Modelshort and previous work on multigraph belief propagation. Experiments on six benchmark datasets show that \Modelshort consistently outperforms strong baseline methods, especially when the original graphs are severely incomplete or the labeled instances for model training are highly sparse\footnote{Our code is available at https://github.com/Maysir/GRCN}. 

\end{abstract}
\section{Introduction}

Graph Convolutional Networks (GCNs) have received increasing attention in recent years as they are highly effective in 
graph-based node feature induction and belief propagation, and widely applicable to many real-world problems,
including computer vision~\cite{wang2018dynamic,landrieu2018large}, natural language processing~\cite{kipf2016semi,marcheggiani2017encoding}, recommender systems~\cite{monti2017geometric,ying2018graph}, epidemiological forecasting \cite{wu2018deep}, and more. 

However, the power of GCNs has not been fully exploited as 
most of the models assume that the given graph perfectly depicts the ground-truth of the relationship between nodes. Such assumptions are bound to yield sub-optimal results as real-world graphs are usually highly noisy, incomplete (with many missing edges), and not necessarily ideal for different downstream tasks. Ignoring these issues is a fundamental weakness of many existing GCN methods.

Recent methods that attempt to modify the original graph can be split into two major streams: 
1) Edge reweighting: GAT \cite{velivckovic2017graph} and GLCN \cite{jiang2019semi} use attention mechanism or feature similarity to reweight the existing edges of the given graph. Since the topological structure of the graph is not changed, the model is prone to be affected by noisy data when edges are sparse. 
2) Full graph parameterization: LDS \cite{franceschi2019learning}, on the other hand, allows every possible node pairs in a graph to be parameterized. Although this design is more flexible, the memory cost is intractable for large datasets, since the number of parameters increases quadratically with the number of nodes. Therefore, finding a balance between model expressiveness and memory consumption remains an open challenge.

To enable flexible edge editing while maintaining scalability, we develop a GCN-based graph revision module that performs edge addition and edge reweighting. In each iteration, we calculate an adjacency matrix via GCN-based node embeddings, and select the edges with high confidence to be added. Our method permits a gradient-based training of an end-to-end neural model that can predict unseen edges. Our theoretical analysis demonstrates the effectiveness of our model from the perspective of multigraph \cite{balakrishnan1997graph}, which allows more than one edges from different sources between a pair of vertices. To the best of our knowledge, we are the first to reveal the connection between graph convolutional networks and multigraph propagation. 
Our contributions can be summarized as follows:

\begin{itemize}
    \item We introduce a novel structure that simultaneously learns both graph revision and node classification through different GCN modules.
    \item Through theoretical analysis, we show our model's advantages in the view of multigraph propagation. 
    \item Comprehensive experiments on six benchmark datasets from different domains show that our proposed model achieves the best or highly competitive results, especially under the scenarios of highly incomplete graphs or sparse training labels. 
\end{itemize}

\section{Background}
We first introduce some basics of graph theory. An undirected graph $G$ can be represented as $(V, E)$ where $V$ denotes the set of vertices and $E$ denotes the set of edges. Let $N$ and $M$ be the number of vertices and edges, respectively. Each graph
can also be represented by an adjacency matrix $A$ of size $N \times N$ where $A_{ij} = 1$ if there is an edge between $v_i$ and $v_j$, and $A_{ij} = 0$ otherwise. We use $A_i$ to denote the $i$-th row of the adjacency matrix. A graph with adjacency matrix $A$ is denoted as $G_A$. Usually each node $i$ has its own feature $x_i \in \mathbb{R}^F$ where $F$ is the feature dimension (for example, if nodes represent documents, the feature can be a bag-of-words vector).  The node feature matrix of the whole graph is denoted as $X \in \mathbb{R}^{N\times F}$. 

Graph convolutional networks generalize the convolution operation on images to graph structure data, performing layer-wise propagation of node features \cite{bruna2013spectral,kipf2016semi,hamilton2017inductive,velivckovic2017graph}. Suppose we are given a graph with adjacency matrix $A$ and node features $H^{(0)} = X$.
An $L$-layer Graph Convolution Network (GCN) \cite{kipf2016semi} conducts the following inductive layer-wise propagation:
\begin{equation}
\begin{split}
H^{(l+1)}=\sigma\left(\widetilde{D}^{-\frac{1}{2}} \widetilde{A} \widetilde{D}^{-\frac{1}{2}} H^{(l)} W^{(l)}\right),
\end{split}
\end{equation}
where $l=0,1,\cdots, L-1$, $\tilde{A} = A + I$ and $\widetilde{D}$ is a diagonal matrix with $D_{ii} = \sum_j \widetilde{A}_{ij}$. $\{ W^{(0)}, \cdots, W^{(L-1)} \}$ are the model parameters and $\sigma(\cdot)$ is the activation function. The node embedding $H^{(L)}$ can be used for downsteam tasks. For node classification, GCN defines the final output as:
\begin{equation}
\begin{split}
\widehat{Y} =\text{softmax}\left(H^{(L)} W^{(L)}\right).
\end{split}
\end{equation}
where $\widehat{Y} \in \mathbb{R}^{N\times C}$ and $C$ denotes the number of classes. We note that in the GCN computation, $A$ is directly used as the underlining graph without any modification. Additionally, in each layer, GCN only updates node representations as a degree-normalized aggregation of neighbor nodes. 

To allow for an adaptive aggregation paradigm, GLCN~\cite{jiang2019semi} learns to reweight the existing edges by node feature embeddings. The reweighted adjacancy matrix $\widetilde{A}$ is calculated by:
\begin{equation}
\begin{split}
\widetilde{A}_{i j}=\frac{A_{i j} \exp \left(\operatorname{ReLU}\left(a^{T}\left|x_{i}P-x_{j}P\right|\right)\right)}{\sum_{k=1}^{n} A_{i k} \exp \left(\operatorname{ReLU}\left(a^{T}\left|x_{i}P-x_{k}P\right|\right)\right)}, 
\end{split}
\label{eq:glcn}
\end{equation}
where $x_i$ denotes the feature vector of node $i$ and $a,P$ are model parameters. 
Another model GAT~\cite{velivckovic2017graph} reweights edges by a layer-wise self-attention across node-neighbor pairs to compute hidden representations. 
For each layer $l$, the reweighted edge is computed by: 
\begin{equation}
\begin{split}
\widetilde{A}^{(l)}_{ij} = \frac{A_{i j} \exp \left(a(W^{(l)} H^{(l)}_i, W^{(l)} H^{(l)}_j)\right)}{\sum_{k=1}^{n} A_{i k} \exp \left(a(W^{(l)} H^{(l)}_i, W^{(l)} H^{(l)}_k)\right)} 
\end{split}
\label{eq:gat}
\end{equation}
where $a(\cdot, \cdot)$ is a shared attention function to compute the attention coefficients. 
Compared with GLCN, GAT uses different layer-wise maskings to allow for more flexible representation. However, neither of the methods has the ability to add edges since the revised edge $\widetilde{A}_{ij}$ or $\widetilde{A}^{(l)}_{ij} \neq 0$ only if the original edge $A_{i j} \neq 0$. 

In order to add new edges into the original graph, LDS~\cite{franceschi2019learning} makes the entire adjacency matrix parameterizable. Then it jointly learns the graph structure $\theta$ and the GCN parameters $W$ by approximately solving a bilevel program as follows:
\begin{equation}
\begin{split}
& \min _{\theta \in \overline{\mathcal{H}}_{N}} \mathbb{E}_{A \sim \operatorname{Ber}(\theta)}\left[\zeta_{val}\left(W_{\theta}, A\right)\right], \\ 
\text { such that } & W_{\theta}=\arg \min _{W} \mathbb{E}_{A \sim \operatorname{Ber}(\theta)}[\zeta_{train}(W, A)],
\end{split}
\end{equation} 
where $A \sim \operatorname{Ber}(\theta)$ means sampling adjacency matrix $A \in \mathbb{R}^{N\times N}$ from Bernoulli distribution under parameter $\theta \in \mathbb{R}^{N\times N}$. $\overline{\mathcal{H}}_{N}$ is the convex hull of the set of all adjecency matrices for $N$ nodes. $\zeta_{train}$ and $\zeta_{val}$ denote the node classification loss on training and validation data respectively. However, this method can hardly scale to large graphs since the parameter size of $\theta$ is $N^2$ where $N$ is the number of nodes. In the next section, we'll present our method which resolves the issues in previous work.

\section{Proposed Method}

\subsection{\Model}

\begin{figure}[!tp]
    \centering
    \includegraphics[width=10cm]{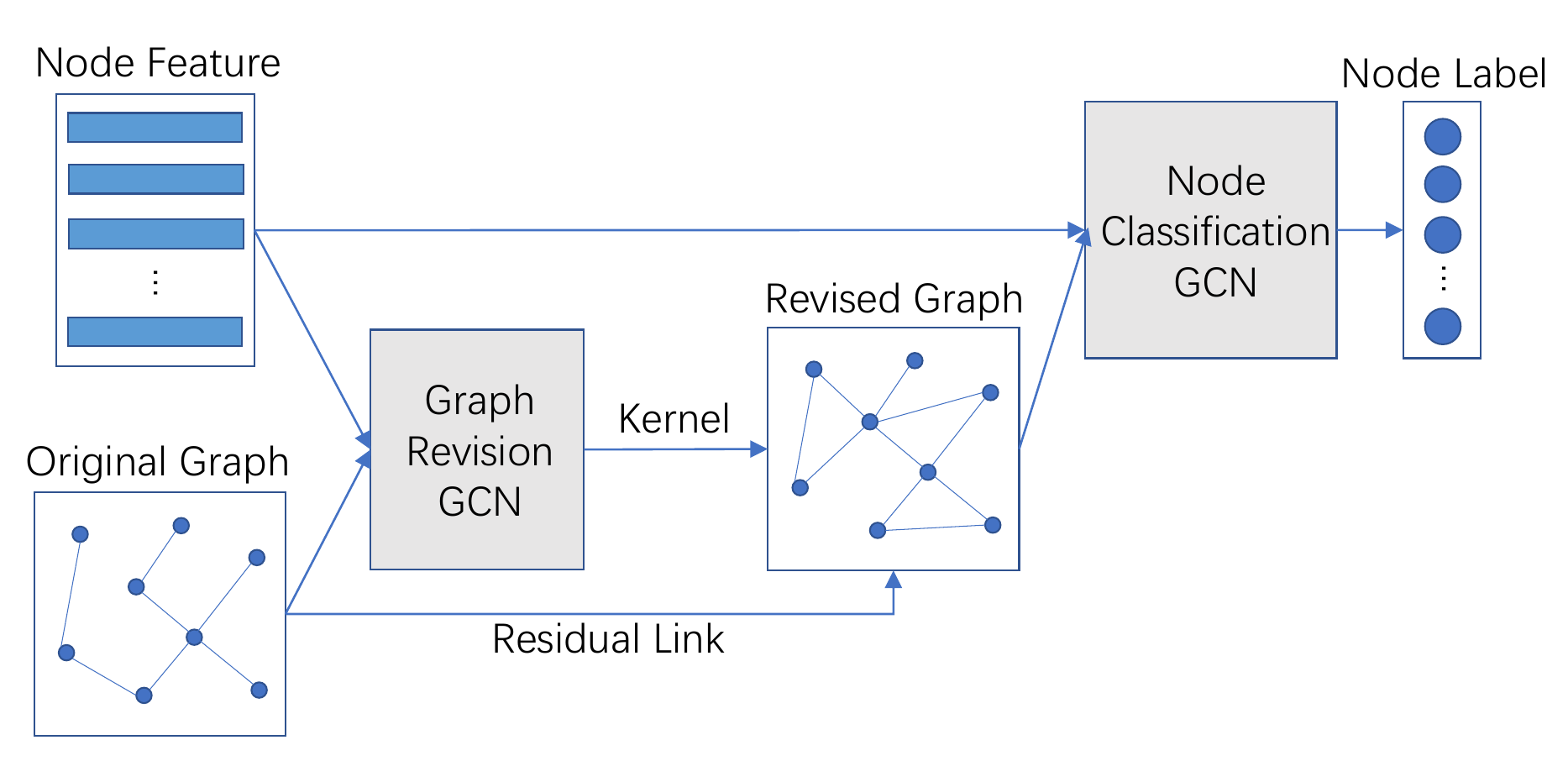}
    \caption{Architecture of the proposed \Modelshort model for semi-supervised node classification. The node classification GCN is enhanced with a revised graph constructed by the graph revision GCN module.}
    \label{fig:model}
\end{figure}
Our \Model (GRCN) contains two modules: a \textit{graph revision} module and a \textit{node classification} module. The graph revision module adjusts the original graph by adding or reweighting edges, and the node classification module performs classification using the revised graph. 
Specifically, in our graph revision module, we choose to use a GCN to combine the node features and the original graph input, as GCNs are effective at fusing data from different sources~\cite{wu2019comprehensive}. 
We first learn the node embedding $Z \in \mathbb{R}^{N \times D}$ as follows:
\begin{equation}
    Z = GCN_g(A, X) \label{eq:GCN_g}
\end{equation}
where $GCN_g$ denotes the graph convolutional network for graph revision, $A$ is the original graph adjacency matrix and $X$ is node feature. Then we calculate a similarity graph $S$ based on node embedding using certain kernel function $k: \mathbb{R}^D \times \mathbb{R}^D \rightarrow \mathbb{R}$:
\begin{equation}
    S_{ij} = k(z_i, z_j).
\end{equation}
The revised adjacency matrix is formed by an elementwise summation of the original adjacency matrix and the calculated similarity matrix: $\widetilde{A} = A+S$. Compared with the graph revision in GAT and GLCN which use entrywise product, we instead adopt the entrywise addition operator ``$+$" in order for new edges to be considered. In this process, the original graph $A$ is revised by the similarity graph $S$, which can insert new edges to $A$ and potentially reweight or delete existing edges in $A$. In practice, we apply a \textit{sparsification} technique on dense matrix $S$ to reduce computational cost and memory usage, which will be introduced in the next section. Then the predicted labels are calculated by:
\begin{equation}
\begin{split}
\widehat{Y} = GCN_c(\widetilde{A}, X)
\end{split}
\end{equation}
where $GCN_c$ denotes the graph convolutional network for the downstream node classification task. Note that to prevent numerical instabilities, renormalization trick~\cite{kipf2016semi} is also applied on the revised adjacency matrix $\widetilde{A}$. Figure \ref{fig:model} provides an illustration of our model. Finally, we use cross-entropy loss as our objective function:
\begin{equation}
\begin{split}
\zeta = -\sum_{i\in \mathcal{Y}_{L}} \sum_{j=1}^C Y_{ij}ln \widehat{Y}_{ij}
\end{split}
\end{equation}
where $\mathcal{Y}_{L}$ is the set of node indices that have labels $Y$ and $C$ is the number of classes. It's worth emphasizing that our model does not need other loss functions to guide the graph revision process. 

Overall, our model can be formulated as:
\begin{equation}
\begin{split}
&\Modelshort(A,X) =GCN_c(\widetilde{A} ,X), \\
&\widetilde{A} = A +  K(GCN_g(A, X)),
\end{split}
\end{equation}
where $K(\cdot)$ is the kernel matrix computed from the node embeddings in Equation~(\ref{eq:GCN_g}). 
In our implementation, we use dot product as kernel function for simplicity, and we use a two-layer GCN \cite{kipf2016semi} in both modules. 
Note that our framework is highly flexible for other kernel functions and graph convolutional networks, which we leave for future exploration.

\subsection{Sparsification}

Since the adjacency matrix $S$ of similarity graph is dense, directly applying it in the classification module is inefficient. Besides, we only want those edges with higher confidence to avoid introducing too much noise. Thus we conduct a $K$-nearest-neighbour (KNN) sparsification on the dense graph: for each node, we keep the edges with top-$K$ prediction scores. The adjacancy matrix of the KNN-sparse graph, denoted as $S^{(K)}$, is computed as:
\begin{equation}
S^{(K)}_{ij} = 
\left\{  
\begin{aligned}
S_{ij} & , & S_{ij} \in topK(S_i), \\
0 & , & S_{ij} \notin topK(S_i).
\end{aligned}  
\right.
\end{equation}
where $topK(S_i)$ is the set of top-$K$ values of vector $S_i$. Finally, in order to keep the symmetric property, the output sparse graph $\widehat{S}$ is calculated by:
\begin{equation}
\widehat{S}_{ij} =
\left\{  
\begin{aligned}
\max(S_{ij}^{(K)}, S_{ji}^{(K)})&,& S_{ij}^{(K)}, S_{ij}^{(K)} \geq 0 \\
\min(S_{ij}^{(K)}, S_{ji}^{(K)})&,& S_{ij}^{(K)}, S_{ij}^{(K)} \leq 0
\end{aligned}  
\right. 
\end{equation}

Now since both original graph $A$ and similarity graph $\widehat{S}$ are sparse, efficient matrix multiplication can be applied on both GCNs as in the training time, gradients will only backpropagate through the top-$K$ values. By sparsification, the memory cost of similarity matrix is reduced from $O(N^2)$ to $O(NK)$.

\subsection{Fast-GRCN}

To further reduce the training time, we introduce a faster version of the propose model: Fast-GRCN. Note that GRCN needs to compute the dense adjacency matrix $S$ in every epoch, where the computational time complexity is $O(N^2)$. While in Fast-GRCN, the whole matrix $S$ is only calculated in the first epoch, and then the indices of the non-zero values of the KNN-sparse matrix $S^{(K)}$ are saved. For the remaining epochs, to obtain $S^{(K)}$, we only compute the values of the saved indices while directly setting zeros for other indices, which reduces the time complexity to $O(NK)$. 

The intuition behind is that the top-$K$ important neighbours of each node may remain unchanged, while the weight of edges between them should still be adjusted during training.

\subsection{Theoretical Analysis}
In this section, we show the effectiveness of our model in the view of Multigraph~\cite{balakrishnan1997graph} propagation. The major observation is that for existing methods, the learned function from GCNs can be regarded as a linear combination of limited pre-defined kernels where the flexibility of kernels have a large influence on the final prediction accuracy.

We consider the simplified graph convolution neural network $GCN_s$ for the ease of analysis. That is, we remove feature transformation parameter $W$ and non-linear activation function $\sigma(\cdot)$ as:
\begin{equation}
\begin{split}
GCN_s(A, X) &= A^kX 
\end{split}
\end{equation}
where $k$ is the number of GCN layers. For simplicity we denote $A$ as the adjacency matrix with self-loop after normalization. The final output can be acquired by applying a linear or logistic regression function $f(\cdot)$ on the node embeddings above:
\begin{equation}
\widehat{Y} = f(GCN_s(A, X)) = f(A^kX)    
\end{equation}
where $\widehat{Y}$ denotes the predicted labels of nodes. Then the following theorem shows that under certain conditions, the optimal function $f^*$ can be expressed as a linear combination of kernel functions defined on training samples.
\begin{theorem}[Representer Theorem~\cite{scholkopf2001generalized}]\label{repre} Consider a non-empty set $\mathcal{P}$ and a positive-definite real-valued kernel: $k: \mathcal{P} \times \mathcal{P} \rightarrow \mathbb{R}$ with a corresponding reproducing kernel Hilbert space $H_k$. If given: a. a set of training samples $\{(p_i, y_i) \in \mathcal{P} \times \mathbb{R}| i=1,\cdots, n\}$, b. a strictly monotonically increasing real-valued function $g: [0, \infty) \rightarrow \mathbb{R}$, c. an error function $E:(\mathcal{P} \times \mathbb{R}^2)^n \rightarrow \mathbb{R}\cup \{ \infty \}$, which together define the following regularized empirical risk functional on $H_k$:
$$f \mapsto E\left(\left(p_{1}, y_{1}, f\left(p_{1}\right)\right), \ldots,\left(p_{n}, y_{n}, f\left(p_{n}\right)\right)\right)+g(\|f\|).$$ 

Then, any minimizer of the empirical risk admits a representation of the form:
$$f^{*}(\cdot)=\sum_{i=1}^{n} \alpha_{i} k\left(\cdot, p_{i}\right)$$
where $a_i \in \mathbb{R} \ \forall i =1, \cdots, n $.
\end{theorem}

In our case, $p_i \in \mathbb{R}^D$ is the embedding of node $i$. As shown in the theorem, the final optimized output is the linear combination of certain kernels on node embeddings. We assume the kernel function to be dot product for simplicity, which means $k(p_i, p_j) = p_i^Tp_j$. The corresponding kernel matrix can be written as:
\begin{equation}
\begin{split}
K(GCN_s(A,X)) & = A^k XX^T A^k 
 = A^k B A^k 
\end{split}
\end{equation}
where $B=XX^T$ is the adjacency matrix of graph induced by node features. Now we have two graphs based on the same node set: original graph $G_A$ (associated with adjacency matrix $A$) and feature graph $G_B$ (associated with adjacency matrix $B$). They form a multigraph~\cite{balakrishnan1997graph} where multiple edges is permitted between the same end nodes. Then the random-walk-like matrix $A^kBA^k$ can be regarded as one way to perform graph label/feature propagation on the multigraph. 
Its limitation is obvious: the propagation only happens once on the feature graph $G_B$, which lacks flexibility. However, for our method, we have:
\begin{equation}
\begin{split}
\Modelshort(A,X) = &(A+ K(GCN_s(A,X)))^kX \\
 = &(A+ A^mXX^TA^m)^kX \\
= &(A +  A^m B A^m)^k X ,\\
K(\Modelshort(A,X))  = &(A +  A^m B A^m)^k B \\ & (A +  A^m B A^m)^k,
\end{split} 
\label{eq:grcnmulti}
\end{equation}
where labels/features can propagate multiple times on the feature graph $G_B$. Thus our model is more flexible and more effective especially when the original graph $G_A$ is not reliable or cannot provide enough information for downstream tasks. In Equation~(\ref{eq:grcnmulti}), $A +  A^m B A^m$ can be regarded as a combination of different edges in the multigraph. To reveal the connection between \Modelshort and GLCN~\cite{jiang2019semi}, we first consider the special case of our model that $m=0$: $GRCN(A,X)=(A+B)^kX$. The operator ``$+$" is analogous to the operator $OR$ which incorporates information from both graph $A$ and $B$. While GLCN~\cite{jiang2019semi} takes another combination denoted as $A \circ B$ using Hadamard (entrywise) product ``$\circ$", which can be analogous to $AND$ operation.

We can further extend our model to a layer-wise version for comparison to GAT~\cite{velivckovic2017graph}. More specifically, for the $l$-th layer, we denote the input as $X_l$. The output $X_{l+1}$ is then calculated by:
\begin{equation}
\begin{split}
X_{l+1} = & (A +  K(GCN_s(A, X_l)))X_{l} \\
= & (A +  A^m X_lX_l^T A^m)X_{l} \\
= & (A + A^mB_lA^m)X_{l},
\end{split}    
\end{equation}
where $B_l=X_lX_l^T$. Similar to the analysis mentioned before, if we consider the special case of \Modelshort that $m=0$ and change the edge combination operator from entrywise sum ``$+$" to entrywise product ``$\circ$", we have $X_{l+1} = (A \circ B_l)X_{l}$, which is the key idea behind GAT~\cite{velivckovic2017graph}. Due to the property of entrywise product, the combined edges of both GAT and GLCN are only the reweighted edges of $A$, which becomes ineffective when the original graph $G_A$ is highly sparse. Through the analysis above, we see that our model is more general in combining different edges by varying the value of $m$, and also has more robust combination operator ``$+$" compared to previous methods.


\section{Experiments}


\begin{table}[!tp]
\begin{center}
\small
\begin{tabular}{lrrrr}
\toprule
Dataset  & \#nodes & \#edges & \#feature & \#class \\ \toprule
Cora     & 2708    & 5429    & 1433      & 7       \\
CiteSeer & 3327    & 4732    & 3703      & 6       \\
PubMed   & 19717   & 44338   & 500       & 3      
\\
CoraFull &  19793  &  65311  & 8710  &  70\\
Amazon Computers &   13381 &  245778  &  767 & 10 \\
Coauthor CS &  18333   &  81894  &  6805 & 15 \\ \toprule
\end{tabular}
\end{center}
\caption{Data statistics}
\label{tab:data}
\end{table}

\subsection{Datasets and Baselines}
We use six benchmark datasets for semi-supervised node classification evaluation. Among them, Cora, CiteSeer~\cite{sen2008collective} and PubMed~\cite{namata2012query} are three commonly used datasets. The data split is conducted by two ways. The first is the fixed split originating from~\cite{yang2016revisiting}. In the second way, we conduct 10 random splits while keeping the same number of labels for training, validation and testing as previous work. This provides a more robust comparison of the model performance. To further test the scalability of our model, we utilize three other datasets: Cora-Full~\cite{bojchevski2018deep}, Amazon-Computers and Coauthor CS~\cite{shchur2018pitfalls}. The first is an extended version of Cora, while the second and the third are co-purchase and co-authorship graphs respectively. On these three datasets, we follow the previous work~\cite{shchur2018pitfalls} and take 20 labels of each classes for training, 30 for validation, and the rest for testing. We also delete the classes with less than 50 labels to make sure each class contains enough instances. The data statistics are shown in Table \ref{tab:data}. 

We compare the effectiveness of our \Modelshort model with several baselines, where the first two models are vanilla graph convolutional networks without any graph revision: GCN~\cite{kipf2016semi}, SGC~\cite{wu2019simplifying}, GAT~\cite{velivckovic2017graph}, LDS~\cite{franceschi2019learning}, and GLCN~\cite{jiang2019semi}.


\begin{table}[!t]
\centering
\setlength{\tabcolsep}{3mm}{
\begin{tabular}{lccc}
\toprule Models
     & \tabincell{c}{Cora\\(fix. split)} & \tabincell{c}{CiteSeer\\(fix. split)} & \tabincell{c}{PubMed\\(fix. split)} \\ \toprule
GCN  &  $81.4\pm0.5$    &   $70.9\pm0.5$       &  $\mathbf{79.0\pm0.3}$      \\
SGC  &   $81.0\pm0.0$   &  $71.9\pm0.1$        & $78.9\pm0.0^{*}$       \\
GAT  &   $83.2\pm0.7$   &   $72.6\pm 0.6$       &   $78.8\pm0.3$     \\
LDS  &   $84.0\pm0.4^{*}$   &   $\mathbf{74.8\pm0.5}$       &   N/A     \\
GLCN &  $81.8\pm0.6$    &    $70.8 \pm 0.5$      &    $78.8\pm0.4$     \\ \hline
Fast-\Modelshort & $83.6\pm0.4$    &  $72.9\pm0.6$     &  $\mathbf{79.0\pm0.2}$    \\
\Modelshort &   $\mathbf{84.2 \pm 0.4}$  &    $73.6 \pm 0.5^{*}$      &  $\mathbf{79.0\pm0.2}$     \\
\toprule Models
     & \tabincell{c}{Cora\\(rand. splits)} & \tabincell{c}{CiteSeer\\(rand. splits)} & \tabincell{c}{PubMed\\(rand. splits)} \\ \toprule
GCN  &  $81.2\pm1.9$    &   $69.8\pm1.9$       &  $77.7\pm2.9^{*}$      \\
SGC  &   $81.0\pm1.7$   &  $68.9\pm2.0$        & $75.8\pm3.0$       \\
GAT  &   $81.7\pm1.9$   &   $68.8\pm 1.8$       &   $77.7\pm3.2^{*}$     \\
LDS  &   $81.6\pm1.0$   &   $71.0\pm0.9$       &   N/A     \\
GLCN &  $81.4\pm1.9$    &    $69.8 \pm 1.8$      &    $77.2\pm3.2$     \\ \hline
Fast-\Modelshort &   $\mathbf{83.8 \pm 1.6}$  &    $72.3 \pm 1.4^{*}$      &    $77.6 \pm 3.2$   \\
\Modelshort &   $83.7 \pm 1.7^{*}$  &    $\mathbf{72.6 \pm 1.3}$      &    $\mathbf{77.9 \pm 3.2}$   \\ \toprule Models
& Cora-Full & \tabincell{c}{Amazon\\Computers} & \tabincell{c}{Coauthor\\CS} \\ \toprule
GCN  &  $\mathbf{60.3\pm0.7}$    &   $81.9\pm1.7$       &   $\mathbf{91.3\pm0.3}$      \\
SGC  &  $59.1\pm 0.7$    &   $81.8 \pm 2.3$       &     $\mathbf{91.3\pm0.2}$    \\
GAT  &  $59.9\pm 0.6$    &    $81.8\pm2.0$      &   $89.5\pm 0.5$     \\
LDS  &   N/A   &    N/A      &  N/A      \\
GLCN &  $59.1\pm 0.7$    &  $80.4\pm1.9$        &  $90.1\pm 0.5$      \\  \hline
Fast-\Modelshort &   $60.2\pm0.5^{*}$   &    $83.5\pm1.6^{*}$      &
$91.2\pm0.4^{*}$ \\
\Modelshort &   $\mathbf{60.3\pm0.4}$   &    $\mathbf{83.7\pm1.8}$      & $\mathbf{91.3\pm0.3}$ \\\toprule
\end{tabular}
}
\caption{\label{tab:main}Mean test classification accuracy and standard deviation in percent averaged for all models and all datasets. For each dataset, the highest accuracy score is marked in \textbf{bold}, and the second highest score is marked using $*$. N/A stands for the datasets that couldn't be processed by the full-batch version because of GPU RAM limitations.}
\end{table}

\subsection{Implementation Details}
Transductive setting is used for node classification on all the datasets. We train \Modelshort for $300$ epochs using Adam~\cite{kingma2014adam} and select the model with highest validation accuracy for test. We set learning rate as \num{1e-3} for graph refinement module and \num{5e-3} for label prediction module. Weight decay and sparsification parameter $K$ are tuned by grid search on validation set, with the search space $[$\num{1e-4}, \num{5e-4}, \num{1e-3}, \num{5e-3}, \num{1e-2}, \num{5e-2}$]$ and $[5, 10, 20, 30, 50, 100, 200]$ respectively. Our code is based on Pytorch~\cite{paszke2017automatic} and one geometric deep learning extension library~\cite{fey2019fast}, which provides implementation for GCN, SGC and GAT. For LDS, the results were obtained using the publicly available code. Since an implementation for GLCN was not available, we report the results based on our own implementation of the original paper.

\subsection{Main Results}
Table \ref{tab:main} shows the mean accuracy and the corresponding standard deviation for all models across the 6 datasets averaged over 10 different runs\footnote{Note that the performance difference of baseline models between fixed split and random split is also observed in previous work~\cite{shchur2018pitfalls}, where they show that different splits can lead to different ranking of models, and suggest multiple random splits as a better choice. Our later experiments are based on the random split setting.}. We see that our proposed models achieve the best or highly competitive results for all the datasets. The effectiveness of our model over the other baselines 
demonstrates that taking the original graph as input for GCN is not optimal for graph propagation in semi-supervised classification. It's also worth noting that Fast-GRCN, with much lower computational time complexity, performs nearly as well as \Modelshort.

\begin{figure}[!t]
    \centering
    \subfigure[Results on Cora]{
    \includegraphics[width=5cm]{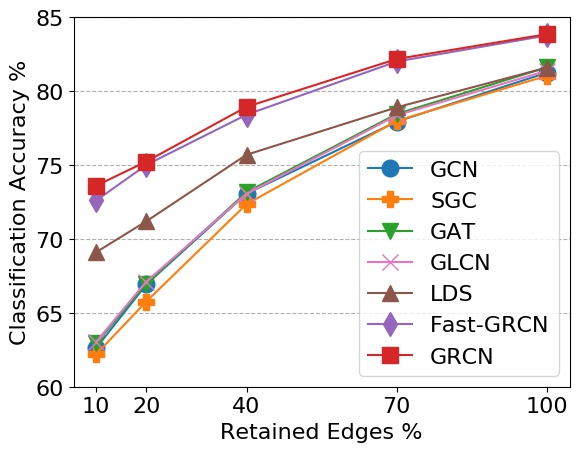}
    }
    \subfigure[Results on CiteSeer]{
    \includegraphics[width=5cm]{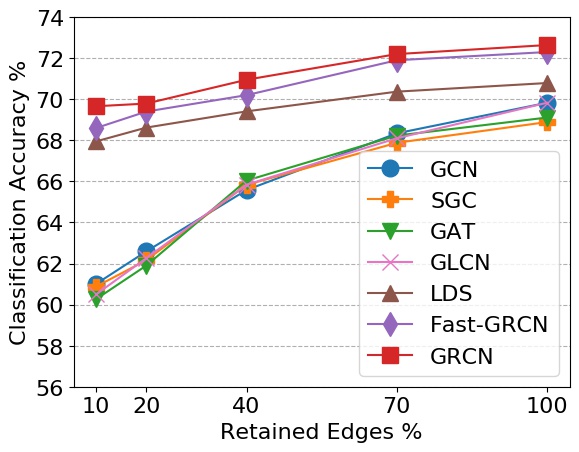}
    }\\
    \subfigure[Results on PubMed]{
    \includegraphics[width=5cm]{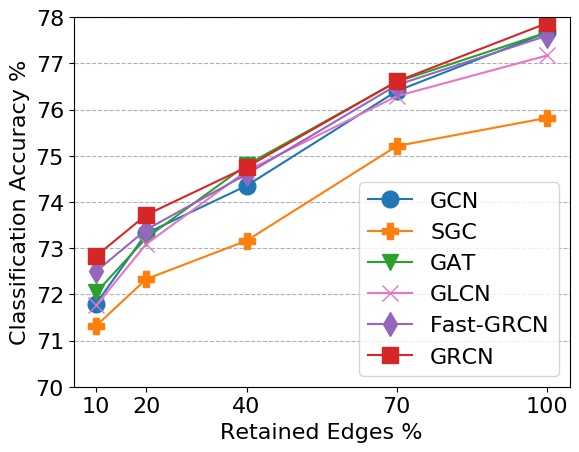}
    }
    \subfigure[Results on Cora-Full]{
    \includegraphics[width=5cm]{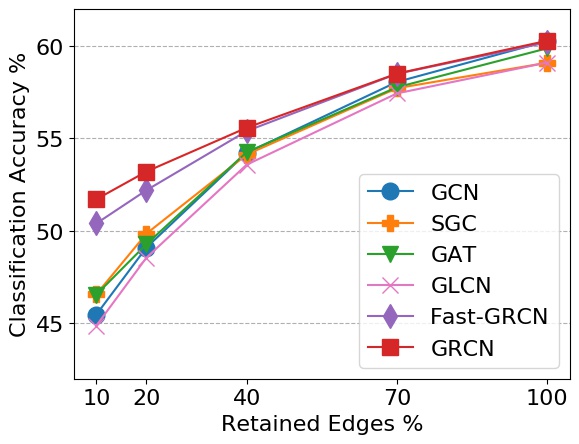}
    }\\
    \subfigure[Results on Amazon Computers]{
    \includegraphics[width=5cm]{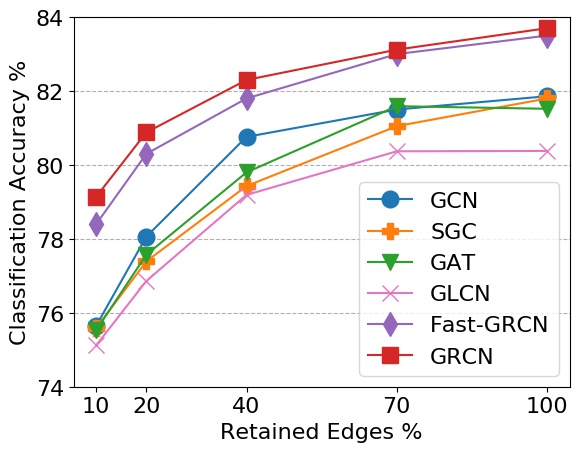}
    }
    \subfigure[Results on Coauthor CS]{
    \includegraphics[width=5cm]{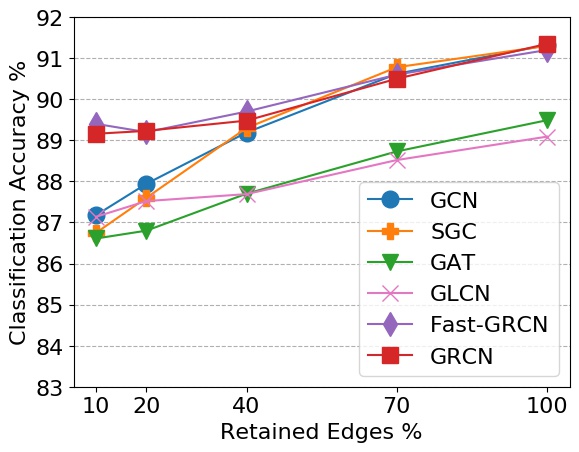}
    }
    \caption{Mean test classification accuracy on all the datasets under different ratios of retained edges over 10 different runs.}
    \label{fig:edge}
\end{figure}

To further test the superiority of our model, we consider the edge-sparse scenario when a certain fraction of edges in the given graph is randomly removed. Given an edge retaining ratio, we randomly sample the retained edges 10 times and report the mean classification accuracy and standard deviation. For the Cora, CiteSeer and PubMed dataset, we conduct experiments under random data split. Figure~\ref{fig:edge} shows the results under different ratios of retained edges. There are several observations from this figure. First, \Modelshort and Fast-\Modelshort achieve notable improvements on almost all the datasets, especially when edge retaining ratio is low. For instance, when edge retaining ratio is $10\%$, \Modelshort outperforms the second best baseline model by $6.5\%, 2.5\%, 1.1\%, 11.0\%, 4.6\%, 2.3\%$ on each dataset. Second, the GAT and GLCN models which reweight the existing edges do not perform well, indicating that such a reweighting mechanism is not enough when the original graph is highly incomplete. Third, our method also outperforms the over-parameterized model LDS in Cora and CiteSeer because of our restrained edge editing procedure. Though LDS achieves better performances than other baseline methods in these two datasets, its inability to scale prevents us from testing it on four of the larger datasets.

\begin{table}[!t]
\centering
\setlength{\tabcolsep}{3.0mm}{
\begin{tabular}{lccc}
\toprule
  Cora-Full   & 5 labels & 10 labels & 15 labels \\ \toprule
GCN  &   $31.3\pm1.5$   &    $41.1\pm1.3$      &  $46.0\pm1.1$    \\
SGC  &  $31.5\pm2.1$   &  $42.0\pm1.5$       &   $46.8\pm1.3$     \\
GAT  &  $32.5\pm2.1$    &    $41.2\pm1.4$      & $45.5\pm1.2$     \\
GLCN &  $30.9\pm1.9$    &  $41.0\pm0.6$        &  $45.0\pm0.9$     \\ \hline
Fast-\Modelshort & $41.2\pm1.2^{*}$ & $47.7\pm0.8^{*}$  & $50.6\pm1.0^{*}$\\
\Modelshort & $\mathbf{42.3\pm0.8}$    &   $\mathbf{48.2\pm0.7}$       & $\mathbf{51.8\pm0.6}$    \\ \toprule
\tabincell{l}{Amazon Computers}  & 5 labels & 10 labels & 15 labels  \\ \toprule
GCN  &  $70.5\pm3.3$    &   $74.6\pm2.3$       &  $77.2\pm2.2$    \\
SGC  &  $67.2\pm5.0$   &    $74.6\pm4.6$     &   $77.1\pm1.6$     \\
GAT  &  $64.6\pm8.9$    &   $72.5\pm4.5$       & $74.2\pm2.7$     \\
GLCN &   $66.9\pm7.1$   &  $73.8\pm3.6$        & $75.8\pm2.2$      \\ \hline
Fast-\Modelshort & $74.1\pm1.9^{*}$ & $78.6\pm2.1^{*}$ &  $79.8\pm1.5^{*}$\\
\Modelshort &   $\mathbf{75.3\pm1.2}$  &    $\mathbf{79.1\pm1.9}$      &   $\mathbf{79.9\pm1.6}$  \\ \toprule
Coauthor CS  & 5 labels & 10 labels & 15 labels \\ \toprule
GCN  &  $82.2\pm1.5$    &   $86.1\pm0.5$       &  $87.1\pm0.9$    \\
SGC  &   $81.5\pm1.6$  &   $85.7\pm0.9$      &  $86.7\pm0.9$      \\
GAT  &  $80.7\pm1.1$    &  $84.8\pm1.0$        &  $86.0\pm0.7$    \\
GLCN &  $82.7\pm0.7$    &   $85.7\pm0.6$       & $ 87.0\pm0.8$      \\ \hline
Fast-\Modelshort & $\mathbf{86.2\pm2.1}$ & $\mathbf{88.2\pm0.7}$ & $\mathbf{88.5\pm0.6}$ \\
\Modelshort &  $86.1\pm0.7^{*}$   &   $87.9\pm0.4^{*}$       & $88.2\pm0.5^{*}$    \\ \toprule
\end{tabular}
\caption{Mean test classification accuracy and standard deviation on Cora-Full, Amazon Computers and Coauthor CS datasets under different number of training labels for each class. The edge retaining ratio is $20\%$ for all the results.The highest accuracy score is marked in \textbf{bold}, and the second highest score is marked using $*$.}
\label{tab:sparse_label}
}
\end{table}

\subsection{Robustness on Training Labels}
We also show that the gains achieved by our model are very robust to the reduction in the number of training labels for each class, denoted by $T$. We compare all the models on the Cora-Full, Amazon Computers and Coauthor CS datasets and fix the edge sampling ratio to $20\%$. We reduce $T$ from $15$ to $5$ and report the results in Table~\ref{tab:sparse_label}. While containing more parameters than vanilla GCN, our model still outperforms others. Moreover, it wins by a larger margin when $T$ is smaller. This demonstrates our model's capability to handle tasks with sparse training labels.


\subsection{Hyperparameter Analysis}

\begin{figure}[!t]
    \centering
    \includegraphics[width=6cm]{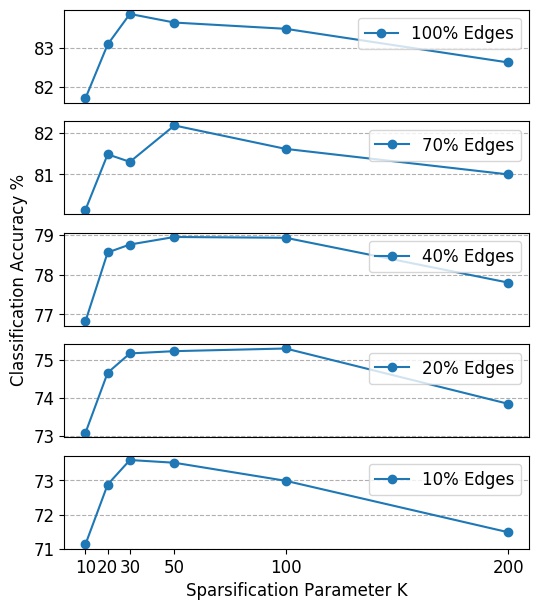}
    \caption{Results of \Modelshort under different settings of sparsification parameter $K$ on Cora dataset, with different edge retaining ratios.}
    \label{fig:K}
\end{figure}

 We investigate the influence of the hyperparameter $K$ in this section. Recall that after calculating the similarity graph in \Modelshort, we use $K$-nearest-neighbour specification to generate a sparse graph out of the dense graph. This is not only benificial to efficiency, but also important for effectiveness. Figure \ref{fig:K} shows the results of node classification accuracy vs. sampling ratio on Cora dataset, where we vary the edge sampling ratio from $10\%$ to $100\%$ and change $K$ from $5$ to $200$. From this figure, increasing the value of $K$ helps improve the classification accuracy at the initial stage. However, after reaching a peak, further increasing $K$ lowers the model performance. We conjecture that this is because a larger $K$ will introduce too much noise and thus lower the quality of the revised graph.

\subsection{Ablation Study}

To further examine the effectiveness of our GCN-based graph revision module, we conduct an ablation study by testing four different simplifications of the graph revision module:

\begin{itemize}
    \item Truncated SVD Reconstruction~\cite{golub1971singular}: $\widetilde{A} = \text{SVD}_{k}(A)$
    \item Feature-Only (FO): $\widetilde{A} = K(X)$
    \item Feature plus Graph (FG): $\widetilde{A} = A + K(X)$
    \item Random Walk Feature plus Graph (RWFG): $\widetilde{A} = A + K(A^2X)$
\end{itemize}
where SVD and FO only use the original graph structure or node features to construct the graph. They are followed by the FG method, which adds the original graph to the feature similarity graph used in FO. Our model is most closely related to the third method, RWFG, which constructs the feature graph with similarity of node features via graph propagation, but without feature learning.

\begin{figure}[ht]
    \centering
    \includegraphics[width=5.5cm]{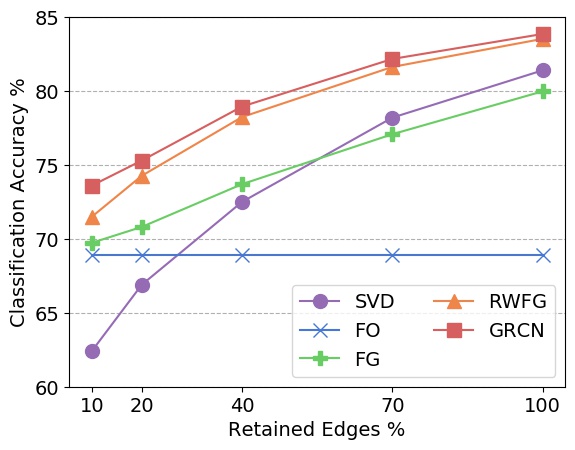}
    \caption{Results of \Modelshort and its simplified versions on Cora dataset with different ratios of retained edges}
    \label{fig:ablation}
\end{figure}

We conduct the ablation experiment on Cora dataset with different edge retaining ratios and report the results in Figure \ref{fig:ablation}. The performance of SVD method indicates that simply smoothing the original graph is insufficient. The comparison between FO and FG shows that adding original graph as residual links is helpful for all edge retaing ratios, especially when there are more known edges in the graph. Examining the results of FG and RWFG, we can also observe a large improvement brought by graph propagation on features. Finally, the performance of our model and RWFG underscores the importance of feature learning, especially in the cases of low edge retraining ratio.

\section{Related Work}

\subsection{Graph Convolutional Network}
Graph Convolution Networks (GCNs) were first introduced in the work by \cite{bruna2013spectral}, with subsequent development and improvements from \cite{henaff2015deep}. Overall, GCNs can be categorized into two categories: spectral convolution and spatial convolution. The spectral convolution operates on the spectral representation of graphs defined in the Fourier domain by the eigen-decomposition of graph Laplacian~\cite{defferrard2016convolutional,kipf2016semi}. 
The spatial convolution operates directly on the graph to aggregate groups of spatially close neighbors~\cite{atwood2016diffusion,hamilton2017inductive}. 
GraphSage \cite{hamilton2017inductive} computes node representations by sampling a fixed-size number of adjacent nodes. 
Besides those methods that are directly applied to an existing graph, GAT \cite{velivckovic2017graph}, GLCN \cite{jiang2019semi} use attention mechanism or feature similarity to reweight the original graph for better GCN performance, while LDS \cite{franceschi2019learning} reconstructs the entire graph via a bilevel optimization. Although our work is related to these methods, we develop a different strategy for graph revision that maintains both efficiency and high flexibility.


\subsection{Link prediction}
Link prediction aims at identifying missing links, or links that are likely to be formed in a given network. 
Previous line of work uses heuristic methods based on local neighborhood structure of nodes, including first-order heuristics by common neighbors and preferential attachment \cite{barabasi1999emergence}, second-order heuristics by Adamic-Adar and resource allocation \cite{zhou2009predicting}, or high-order heuristics by PageRank \cite{brin1998anatomy}. To loose the strong assumptions of heuristic method, a number of neural network based methods~\cite{grover2016node2vec,zhang2018link} are proposed, which are capable to learn general structure features. The problem we study in this paper is related to link prediction since we try to revise the graph by adding or reweighting edges. However, instead of treating link prediction as an objective, our work focus on improving node classification by feeding the revised graph into GCNs.

\section{Conclusion}

This paper presents Graph-Revised Convolutional Network, a novel framework for incorporating graph revision into graph convolution networks. We show both theoretically and experimentally that the proposed way of graph revision can significantly enhance the prediction accuracy for downstream tasks. \Modelshort overcomes two main drawbacks in previous approaches to graph revision, which either employ over-parameterized models and consequently face scaling issues, or fail to consider missing edges.
In our experiments with node classification tasks, the performance of \Modelshort stands out in particular when the input graphs are highly incomplete or if the labeled training instances are very sparse. 

In the future, we plan to explore \Modelshort in a broader range of prediction tasks, such as knowledge base completion, epidemiological forecasting and aircraft anomaly detection based on sensor network data.

\section{Acknowledgments}
We thank the reviewers for their helpful comments. This work is supported in part by the National Science Foundation (NSF) under grant IIS-1546329.

%
%
%
\bibliographystyle{splncs04}

\end{document}